\documentclass{article} 
\usepackage{iclr2020_conference,times}

\usepackage[export]{adjustbox} 
\usepackage{amsmath}
\usepackage{amssymb}
\usepackage{authblk}
\usepackage{booktabs}
\usepackage{csvsimple}
\usepackage{graphicx}
\usepackage{hyperref}
\usepackage{numprint}
\usepackage{url}
\usepackage{siunitx}
\usepackage{subfig}
\usepackage[breakwords]{truncate}
\usepackage{wrapfig}

\usepackage{amsmath,amsfonts,bm}

\def\eqref#1{equation~\ref{#1}}

\def\1{\bm{1}}

\DeclareMathAlphabet{\mathsfit}{\encodingdefault}{\sfdefault}{m}{sl}
\SetMathAlphabet{\mathsfit}{bold}{\encodingdefault}{\sfdefault}{bx}{n}

\def\cB{{\mathcal{B}}}

\def\cE{{\mathcal{E}}}

\def\cR{{\mathcal{R}}}

\def\cX{{\mathcal{X}}}

\newcommand{\eqdef}{\triangleq}

\newcommand{\paren}[1]{\kern-0.5ex\left( #1 \right)}

\newcommand{\sqparen}[1]{\left[ #1 \right]}

\newcommand{\Ex}[2][]{
\ifx\hfuzz#1\hfuzz 
\mathbb{E}\sqparen{#2}
\else
\mathbb{E}_{#1}\sqparen{#2}
\fi
}

\newcommand{\bx}{\mathbf{x}}
\newcommand{\bz}{\mathbf{z}}

\newcommand{\by}{\mathbf{y}}

\newcommand{\cF}{\mathcal{F}}

\usepackage{ifthen}
\newboolean{include_RF}
\setboolean{include_RF}{false}

\title{HydroNets: Leveraging River Structure for Hydrologic Modeling}

\author[1]{Zach Moshe}
\author[1]{Asher Metzger}
\author[1,2]{Gal Elidan}
\author[4]{Frederik Kratzert}
\author[1]{Sella Nevo}
\author[1,3]{Ran El-Yaniv}
\affil[1]{Google Research}
\affil[2]{The Hebrew University of Jerusalem}
\affil[3]{Technion - Israel Institute of Technology}
\affil[4]{LIT AI Lab \& Institute for Machine Learning, Johannes Kepler University Linz}

\iclrfinalcopy 

\begin{document}

\maketitle

\begin{abstract}
Accurate and scalable hydrologic models are essential building blocks of several important applications, from water resource management to timely flood warnings. However, as the climate changes, precipitation and rainfall-runoff pattern variations become more extreme, and accurate training data that can account for the resulting distributional shifts become more scarce.
In this work we present a novel family of hydrologic models, called HydroNets, which leverages river network structure. 
HydroNets are deep neural network models designed to exploit both basin specific rainfall-runoff signals, and upstream network dynamics, which can lead to improved predictions at longer horizons.
The injection of the river structure prior knowledge reduces sample complexity and allows for scalable and more accurate hydrologic modeling even with only a few years of data.
We present an empirical study over two large basins in India that convincingly support the proposed model and its advantages.
\end{abstract}

\section{Introduction}
\label{sec:intro}

Prior knowledge plays an important role in machine learning and AI. On one extreme of the spectrum there are expert systems, which exclusively rely on domain expertise encoded into a model. On the other extreme there are general purpose methods, which are exclusively data-driven. 
In the context of hydrologic modeling, conceptual models such as the Sacramento Soil Moisture Accounting Model (SAC-SMA) \citep{burnash1973generalized}, are analogues to expert systems and require explicit functional modeling of water volume flow.
Instances of agnostic methods 
have recently been presented by \cite{kratzert2018rainfall,kratzert2019towards}
and by \cite{shalev2019accurate}, showing that general purpose deep recurrent neural networks
can achieve state-of-the-art hydrologic forecasts at scale.

Global climate changes resulting in new weather patterns 
can cause rapid distributional shifts that make learned models
irrelevant.
In particular, relevant or recent data is scarce by definition and learning from such data can lead to substantial overfitting. Our goal in this work is to incorporate useful prior knowledge into machine learned hydrologic models so as to overcome this obstacle.

We present \emph{HydroNets}, a family of deep neural network models designed for hydrologic forecasting. 
HydroNets leverages the prior knowledge of the sub-basins' structure of a hydrologic region. HydroNets also enforce some weight sharing between sub-basins, resulting in a shared model and basin-specific models that correspond to the general-physical hydrologic modeling which is shared among basins vs. the basin-specific modeling that account for basin properties.
The proposed architecture is modular, thus making it convenient to understand and improve.
We present experimental results over two regions in India which convincingly show that the proposed model utilizes learning examples from the whole region, avoids overfitting, and performs better when training data is scarce.

\section{Problem Setting}
\label{sec:setting}
We define a \emph{hydrologic region} $\cR$ to be a directed graph, $\cR=(\cB,\cE)$, where each node in the node set,
$\cB = \{b_1, \ldots, b_n \}$, represents a basin and each directed edge, 
$b_i \to b_j \in \cE$, indicates that $b_i$ is a direct sub-basin of $b_j$.  An edge direction corresponds to water flow from a sub-basin to its containing basin, 
and whenever an edge $b_i \to b_j$ exists we say that $b_i$ is a \emph{source} of 
$b_j$ (there can be multiple sources), and $b_j$ is the \emph{downstream} node of $b_i$.
A basin whose out-degree is zero is called a \emph{region drain}. A basin without sources
(whose in-degree is zero) is called a \emph{region source}. 
For each basin $b \in B$ we denote by $S(b) \subseteq B$ the set of sources of $b$.

\begin{wrapfigure}[10]{r}{0pt}
    \includegraphics[width=0.4\textwidth]{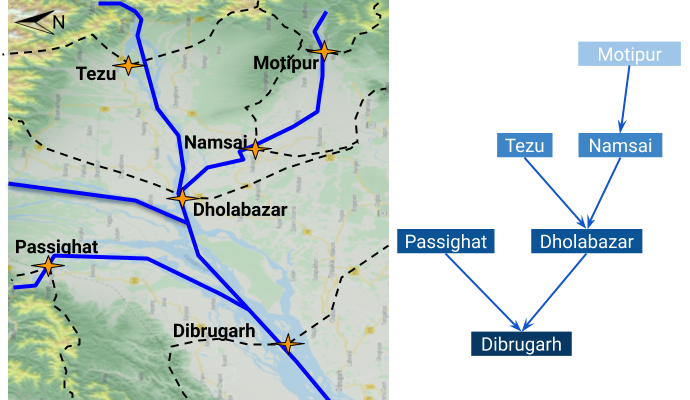}
    \caption{Hydrologic region example}
    \label{fig:NetworkExample}
\end{wrapfigure}
Naturally, due to the topological properties of rivers, sub-basins' structure span an ``inverted tree``. Figure-\ref{fig:NetworkExample} shows an example for such a hydrologic region.

For each basin $b_i$ we consider a sequence of its \emph{temporal features}, 
$\cX_i^{1:t} \eqdef \bx_i^{(1)}, \bx_i^{(2)}, \ldots, \bx_i^{(t)},$
where $\bx_i^{(t)}$ is the feature vector of time $t$, which can include features such as \emph{precipitation}, \emph{temperature}, past readings of the gauge itself, and so on.
For each basin, we also include a vector, $\bz_i$,  of \emph{static features}, which are specific to basin $i$ and are fixed through time. Such feature can include soil type, elevation, etc..

For each basin $b_i$, let $\by_i^{1:t}$
be its \emph{target label} sequence. Typically, target labels are \emph{water-levels} or \emph{discharges} (i.e., the volumetric flow rate of water).
Given a desired prediction horizon, $h$ (say, two days), 
the task is to create model 
$F_{\cR}$ for region $\cR$ that accurately forecasts 
the target labels of all basins
at horizon $h$
from a past window of inputs of length $T$ (e.g., a month),
$
F(\cX_1^{(t-T:t)},\ldots,\cX_n^{(t-T:t)}, \bz_1,\ldots, \bz_n) \rightarrow (y_1^{t+h},\ldots,y_n^{t+h}).
$
In hydrologic forecasting, prediction quality is traditionally measured using the Nash–Sutcliffe efficiency (NSE) \citep{nash1970river},
which is equivalent to the $R^2$ (``variance explained'') of classical statistics. In this work we will also use the $R^2$-persist metric as defined in Appendix-\ref{appendix:defining_r2_persist}.

\begin{wrapfigure}{r}{200pt}
        \vspace{-40pt}
        \begin{tabular}{l}
            \includegraphics[width=0.45\textwidth]{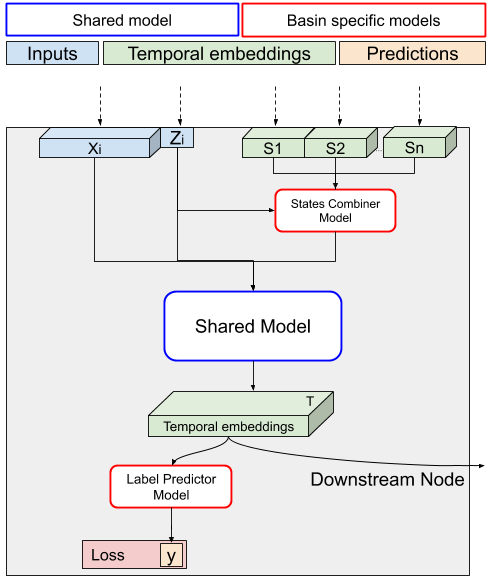} \vspace{0.0in} \\
        $F_i^\text{CMB}\left( \left\{ E_j^{(t-T:t)} \mid j \in S(b_i) \right\}, \bz_i\right) \rightarrow C_i^{(t-T:t)} $  \vspace{0.05in} 
        \\
        $F^\text{SHA}\left(\cX_i^{(t-T:t)}, \bz_i, C_i^{(t-T:t)} \right) \rightarrow E_i^{(t-T:t)}$ \vspace{0.05in} 
        \\
        $F_i^\text{PRD}\left( E_i^{(t-T:t)} \right) \rightarrow l_i^{(t+h)}$

        \end{tabular}
    \vspace{-0.1in}
    \caption{HydroNets Architecture}
    \label{fig:HydroNetsArchitecture}
\end{wrapfigure}

\section{HydroNets}

We propose a novel family of architectures for hydrologic forecasting. Models in
this family, called \emph{HydroNets}, leverage the prior information provided by the river's structure.
Given a hydrologic region $\cR=(\cB,\cE)$, HydroNets spans a computation graph that follows $\cR$ such that for every basin $b_i \in \cB$ in the river graph, the network, $H(\cR) \eqdef (H_1,\ldots, H_n)$, contains a sub-network (also called
a \emph{node})
$H_i$. $H_i$ is connected to $H_j$ iff
($b_i \to b_j) \in \cE$.
Each node $H_i$ is composed of  
three sub-models, two of which are basin-specific 
and the third is shared among all basins. The role of these sub-models 
is explained below.
In each node $b_i$, the shared model outputs a \emph{temporal embedding} vector which encodes self and upstream information for this basin. Additionally, making use of this embedding, a basin-specific model outputs the target label (e.g., water level) at basin $b_i$.
All nodes, other than the region's drain basin, pass their temporal embeddings to their downstream node.
We define $K \eqdef |E^{(t)}_i|$, the size of every embedding vector and consider it as a hyperparameter  of the network.
We now describe each of the three sub-models. Functional forms are given in Figure~\ref{fig:HydroNetsArchitecture}.

\noindent
\textbf{Combiner.}
Each basin $b_i$ receives as input its static features vector,
$\bz_i$, as well as all the temporal embedding of its sources in $S(b_i)$. These inputs are fed to a 
basin-specific sub-model, $F_i^\text{CMB}$, called \emph{combiner}. The output of the combiner is a $T \times K$ matrix, denoted $C_i^{(t-T:t)}$.
The combiner allows each node to handle a different number of sources, and moreover,
allows the node to account for the relative importance of its
sources, which depend on the distances to the sources, 
relative water volume of the sources, and so on.

\noindent
\textbf{Shared Hydrologic Model.} 
The basin-specific output of the combiner at each node, $C_i^{(t-T:t)}$, as well as 
its temporal and static features, 
$(\cX_i^{(t-T:t)}, \bz_i)$,
are fed as inputs to a shared
model $F^\text{SHA}$. This model computes the temporal embeddings for this node and its output is a $T \times K$ matrix $E_i^{(t-T:t)}$.

\noindent
\textbf{Basin-Specific Prediction Model.} 
Based on the temporal embeddings of node $i$, its basin-specific prediction model, denoted $F_i^\text{PRD}$, predicts the target label at time $t+h$.
This allows HydroNets to account for basin-specific behavior. 

Figure~\ref{fig:HydroNetsArchitecture} depicts a single node in the HydroNets architecture. HydroNets recursively builds such nodes by traversing the graph $\cR$ starting with the region drain. The resulting computational graph is thus a tree that matches $\cR$.
The loss function used to optimize our model is a weighted sum over all \emph{MSE} terms between every $l_i$ and its corresponding $y_i$.

\section{Empirical Study}

In the empirical study presented here, we instantiated HydroNets such that all sub-models are linear.
The handling of temporal embedding vectors by each of these sub-models
is done such that the same weights are used on all time steps. 
Throughout our study, we use the following \emph{flat linear} baseline predictor which does not utilize the hydrologic structure (i.e., concatenates the features).
Note that our implementation of HydroNets (and the baseline) 
does not include static features.

\noindent
{\bf The Ganga and Brahmaputra Datasest.} 
The datasets used in our experimental study were constructed 
from two main sources. For precipitation we relied on JAXA's GSMap satellite \citep{ushio2003global}, which generates hourly images of rainfall intensity. 
Water level measurements were taken from the Indian Central Water Commission. 
For this study we constructed two sub-regions from the Brahmaputra and the Ganga rivers.
More detailed maps are in Appendix-\ref{appendix:regions_networks}.

We extracted the polygon describing each basin's geo-spatial location using the HydroSHEDS datasets \citep{lehner2008new, lehner2013global} and calculated the lumped average rain intensity over the basin. 
Every example (i.e., timestamp) in the dataset contains a historical window of length $T$ of the precipitation and past water-levels as the features, and the measurement at $t+h$ as the label.
Our dataset contains five monsoon periods (Jun to Oct) during the years 2014 to 2018. We used the first four years for training and the last one for testing.

\begin{wrapfigure}{r}{0\textwidth}
    \raisebox{5pt}[\dimexpr\height-0.6\baselineskip\relax]{
        \subfloat[Brahmaputra region]{{
            \includegraphics[height=9em,width=10em]{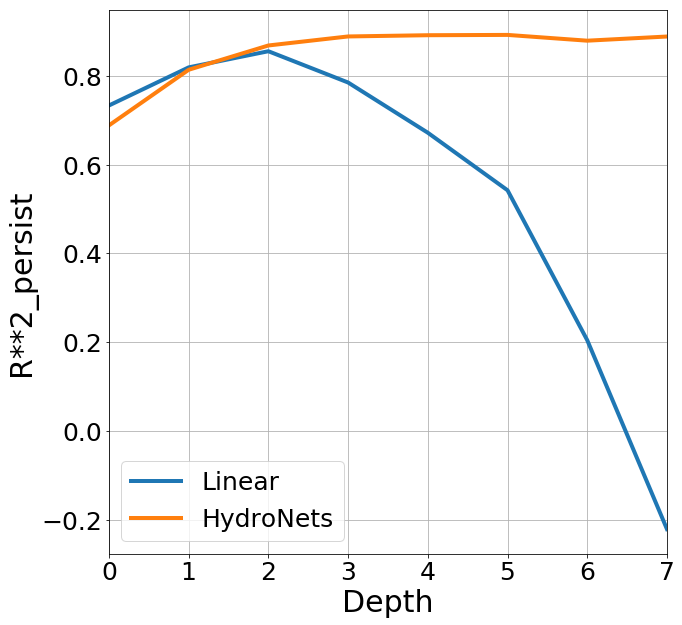}
        }}
        \subfloat[Ganga region]{{
            \includegraphics[height=9em,width=10em]{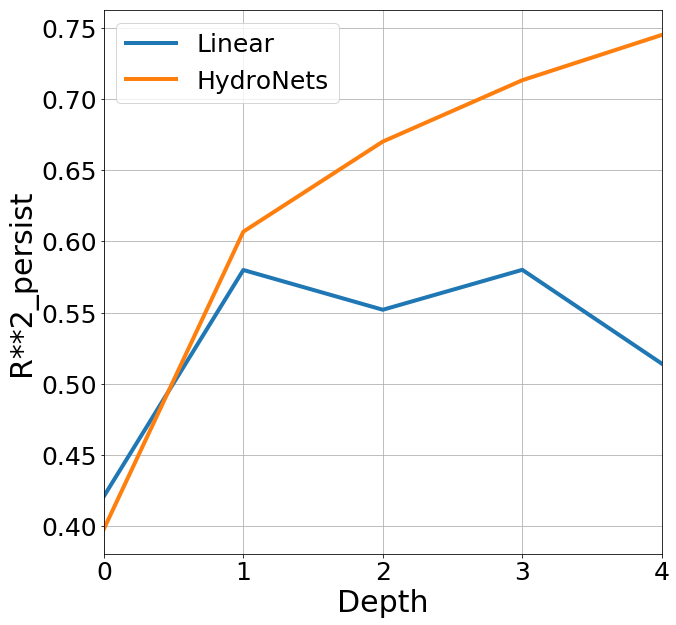}
        }}
    }
    \caption{R-squared on different tree depths}
    \label{fig:DescendantsDepth}
\end{wrapfigure}

\noindent
{\bf Experiment 1: The Value of Depth.} 
We consider the effect of using trees of different depths on prediction at the drain basin of each region.
The $R^2$-persist metric results are presented in Figure~\ref{fig:DescendantsDepth} where we observe that initially both models gain from deepening the tree but while the baseline model starts deteriorating, the HydroNets model keeps leveraging information from deeper branches.
Qualitatively similar results with the standard $R^2$ metric are presented in Appendix-\ref{appendix:r2_results}.

\noindent
{\bf Experiment 2: All Basins Comparison.} 
We examine the performance of HydroNets in each region relative to the flat linear baseline at all sites.
For each basin, we trained a different model where the loss weights were heavily adjusted towards this basin. 
%
%
The flat model was trained for every basin separately with a depth of 2 (following our previous experiment).

\begin{wrapfigure}{r}{0.61\textwidth}
    \raisebox{0pt}[\dimexpr\height-0.6\baselineskip\relax]{
        \subfloat[Brahmaputra region]{{
            \includegraphics[width=0.30\textwidth, valign=t]{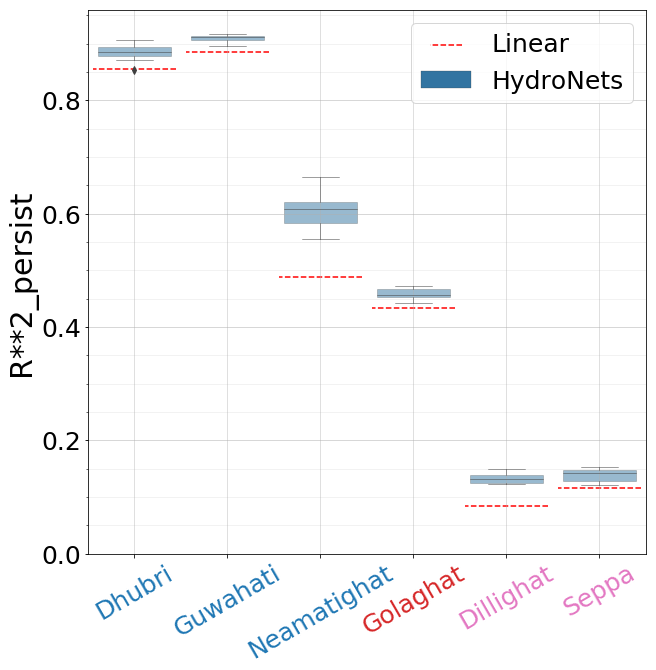} 
        }}
        \subfloat[Ganga region]{{
            \includegraphics[width=0.30\textwidth, valign=t]{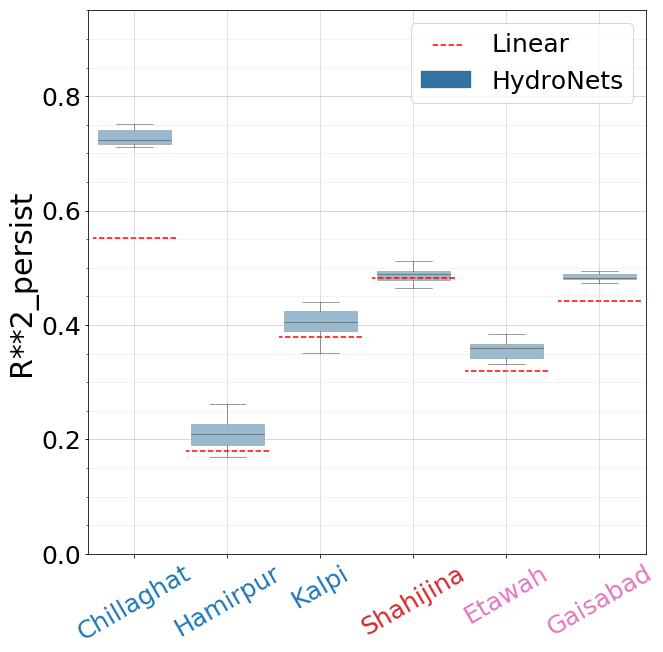}
        }}
    }
    \caption{Persist R-squared on different basins}
    \label{fig:MainComparison}
\end{wrapfigure}

We selected 6 representative basins from each region, 
based on a hydrologic context, where we balanced between large basins where the gauges are located on the main rivers, and smaller upstream basins, which are region sources (leaves) in the region graph.
Figure~\ref{fig:MainComparison} visualizes average results over 10 random initializations and shows that HydroNets outperforms the flat linear baseline in all 6 representative basins. We also see that some basins are harder to predict than the others. This tends to be the case with the more up-stream basins.
Appendix-\ref{appendix:all_basins_results} presents the results for all sub-basins in both regions, where HydroNets outperforms the flat linear model in a large majority of cases, and provides comparable performance in the rest.

\begin{figure}[h]
    \subfloat[Golaghat]{{\includegraphics[width=13em,valign=t]{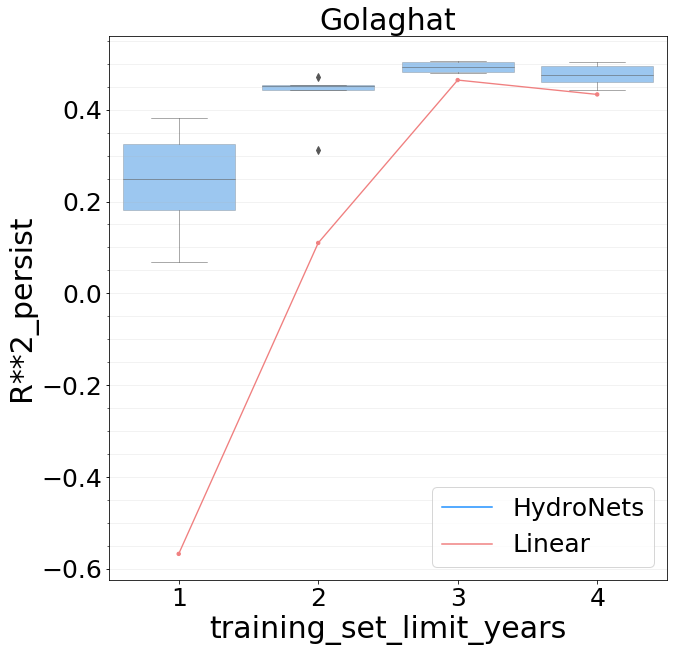} }}
    \subfloat[Neamatighat]{{\includegraphics[width=13em,valign=t]{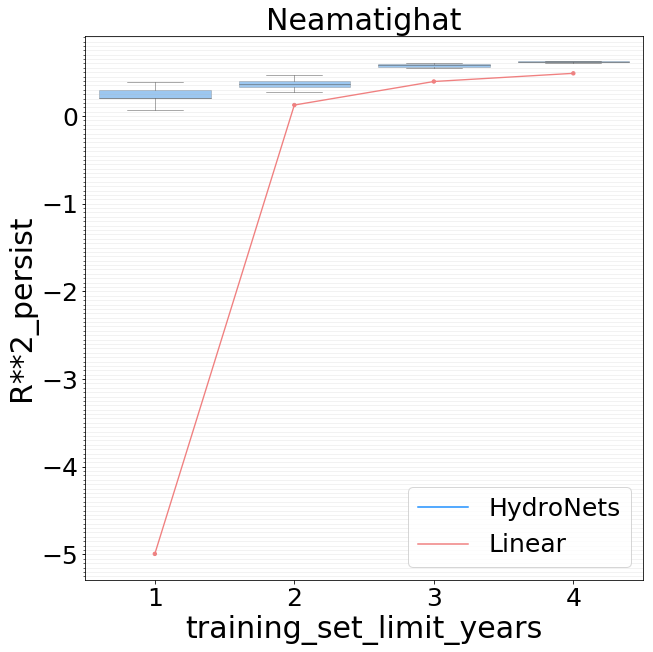} }}
    \subfloat[Chillighat]{{\includegraphics[width=13em,valign=t]{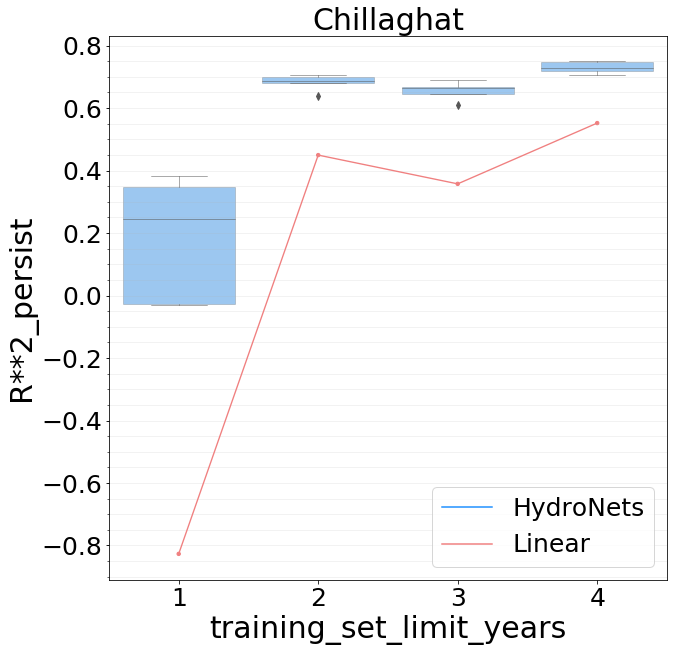} }}
    \vspace{-10pt}
    \caption{$R^2$-persist when increasing number of training years}
    \label{fig:LimitedTrainingset}
\end{figure}

\noindent
{\bf Experiment 3: Learning from fewer samples.} 
We examine the performance achieved using a significantly smaller training set.
Instead of utilizing the entire four years in our datasets, in this study we 
present forecasting performance when using only the last years for training.
In all cases, the test set is fixed to be the fifth year in our datasets.
The results indicate that HydroNets has a substantial and increasing advantage over
the flat linear baseline when the training set becomes smaller.
Figure~\ref{fig:LimitedTrainingset} shows the results for three examples of sub-basins.

\vspace{-10pt}

\vspace{-5pt}
\section{Related Work}

Up to date, most hydrologic models are physical models such as SAC-SMA \citep{burnash1973generalized} and WRF-hydro \citep{salas2018towards}. 
\cite{kratzert2019towards} presented a regional model for hundreds of gauged basins
that strongly depends on basin-specific static features such as area, soil type, etc. and exhibited state-of-the-art streamflow forecasting performance.
While in the work of \cite{kratzert2019towards} the network used static catchment attributes derived from gridded data products, \cite{shalev2019accurate} showed that in the fully gauged setting the same model can be used without static catchment attributes but with a learned site embedding.

\vspace{-5pt}
\section{Concluding Remarks}
We presented HydroNets, a family of architectures for hydrologic modeling. A distinct advantage of the HydroNets architecture 
is that it reduces the sample complexity.
This property enables forecasting in basins where training data is scarce, or when patterns change rapidly, perhaps due to climate change.
HydroNets is a flexible family of models and in this work we only 
considered linear instantiations of its sub-models. Future work may include non-linear and recurrent sub-models, experimenting in other regions, working with discharge as the label and adding static basin features to the implementation.

\newpage

\bibliography{hydronets_bib}
\bibliographystyle{iclr2020_conference}

\newpage
\appendix

\section{Defining the $R^2$-persist metric}
\label{appendix:defining_r2_persist}

In this work we introduce and utilize a performance metric, which we term \emph{$R^2$-persist}. 
Both the standard $R^2$ metric, a.k.a. Nash–Sutcliffe efficiency (NSE) \citep{nash1970river}, and the proposed $R^2$-persist metric have the same general form as a ratio between the mean squared error (MSE) of the model's prediction, relative to the MSE of the predictions of a baseline,
$$
1 - \frac{\text{MSE}(\cF_\cR)}{\text{MSE}(\text{BASELINE})}
$$
In the \emph{NSE} ($R^2$) metric, the baseline is taken to be the average predicted value, while in the $R^2$-persist metric, the baseline is a naive model that always predicts the future using the present reading of the target label measurement  (i.e., $\hat{y}_i^{t+h} = y_i^t$). 

The motivation for introducing this new metric is that in large rivers, the persist baseline is a much stronger
model than the average baseline, which makes the $R^2$-persist a more challenging and meaningful performance 
measure. The values of both metrics are in the interval $(-\infty, 1]$, and near zero values reflect baseline performance.

\section{Detailed maps of the two regions}
\label{appendix:regions_networks}

\begin{figure}[h]
    \centering
    \subfloat[Brahmaputra region]{{\includegraphics[height=15em]{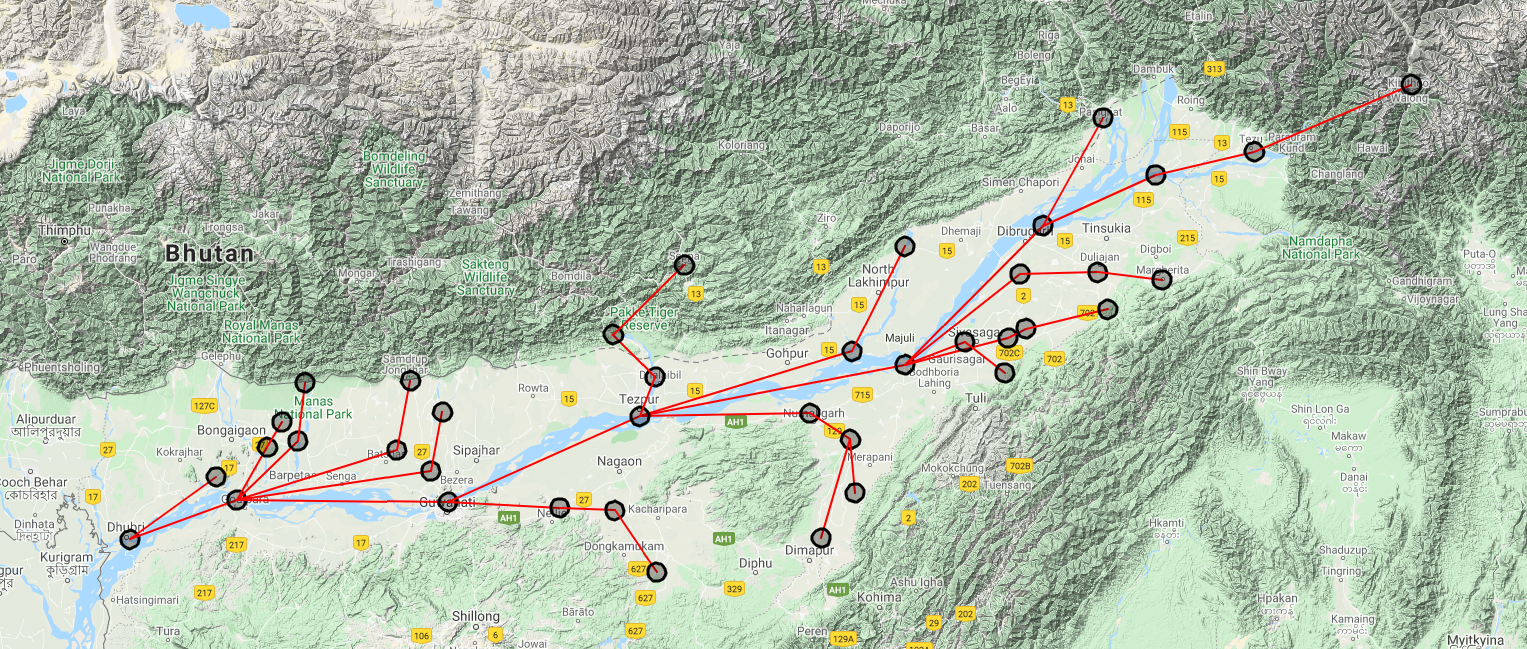} }}
    \qquad
    \subfloat[Ganga region]{{\includegraphics[height=15em]{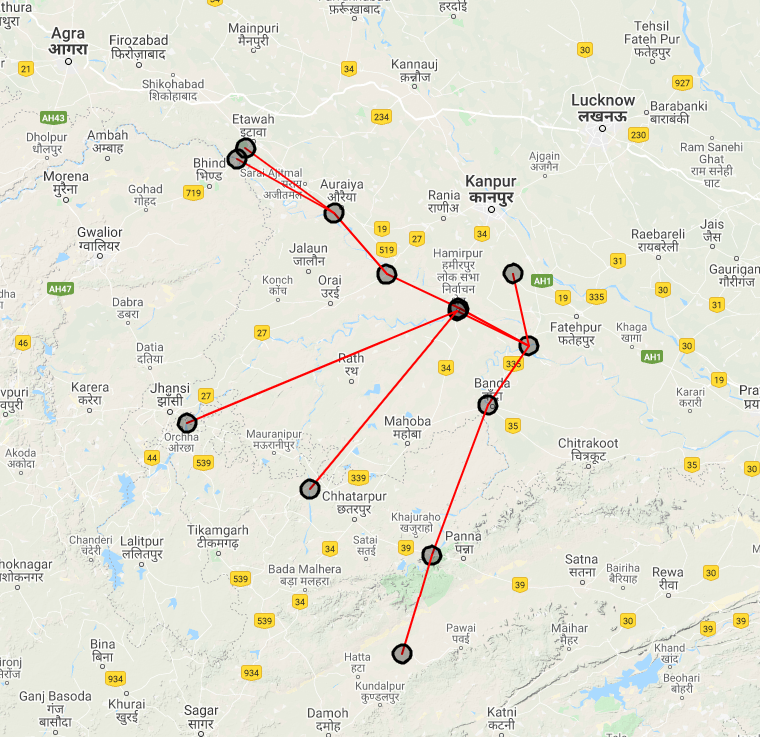} }}
    
    \caption{The Brahmaputra and the Ganga regions}
    \label{fig:RegionsNetworks}
\end{figure}

Figure \ref{fig:RegionsNetworks} shows the two hydrologic regions we used to construct the Brahmaputra 
and Ganga datasets that were used in our empirical study.
Both regions are in India, where
the Brahmaputra region is located in the east part of the Brahmaputra river, and the Ganga region is a small part of the whole Ganga basin, which is located near the city of Lucknow.
\vspace{1em}

\newpage 

\section{Results over all basins}
\label{appendix:all_basins_results}

Table~\ref{table:all_basins_results} shows the results of Experiment 2 for all basins in both regions. The table shows the $R^2$-persist metric.  
For each region we also present a histogram of the \emph{diff} column values. As can be seen, the HydroNets model outperforms the linear model in a large majority of basins. 
\vspace{1em}
\csvstyle{csvStyle}{
    tabular=|l|l|l|l|,
    table head=\hline Basin~Name & \bfseries Linear & \bfseries HydroNets & \bfseries Diff \\\hline\hline,
    late after line=\\\hline,
    head to column names
}

\begin{table}[htb]
    \centering
    \npdecimalsign{.}
    \nprounddigits{3}
    \begin{tabular}{c|c}
        \begin{tabular}[t]{l n{1}{3} n{1}{3} n{1}{3}}
            \multicolumn{4}{c}{Brahmaputra} \\
            \toprule
            {Basin Name} & {Linear} & {HydroNets} & {Diff} \\
            \midrule
Badatighat & 0.5568443865666352 & 0.5657635612645384 & 0.008919174697903265 \\
Behalpur & 0.32078971572396364 & 0.3289656188027492 & 0.008175903078785574 \\
Beki Road bridge & 0.14243647528059078 & 0.17420767157737682 & 0.03177119629678604 \\
Bhalukpong & -0.05450698045963054 & 0.1634540042403937 & 0.21796098470002423 \\
Bihubar & 0.27566067687928986 & 0.2723152605579484 & -0.0033454163213414434 \\
Bokajan & 0.3166888617558683 & 0.32879294190163033 & 0.01210408014576203 \\
Chenimari & 0.7200725865820619 & 0.7011669381339638 & -0.018905648448098056 \\
Chouldhowaghat & 0.255339071731416 & 0.254241935452059 & -0.001097136279357036 \\
Desangpani & 0.37805852442175314 & 0.39995152720354454 & 0.021893002781791404 \\
Dharamtul & 0.5510783029520876 & 0.5618626605554085 & 0.01078435760332086 \\
Dholabazar & -1.3652369270272944 & -1.1181052084940486 & 0.24713171853324578 \\
Dhubri & 0.8551722177250006 & 0.8650109743837727 & 0.009838756658772096 \\
Dibrugarh & -0.44001122603952836 & 0.22348232535742274 & 0.6634935513969511 \\
Dillighat & 0.08493667910130365 & 0.13724150655998368 & 0.05230482745868004 \\
Gelabil & 0.10128728796543685 & 0.12967692153513977 & 0.02838963356970292 \\
Goalpara & 0.8512461899775181 & 0.8605118379156127 & 0.00926564793809459 \\
Golaghat & 0.43325555039895314 & 0.4667733043544595 & 0.03351775395550638 \\
Guwahati & 0.8857147553453816 & 0.9106473936456345 & 0.02493263830025294 \\
Jiabharali    NT Road          X-ing & -0.014185175560048169 & 0.18156865976756809 & 0.19575383532761625 \\
Kampur & 0.5036722171224572 & 0.5375832845667163 & 0.03391106744425909 \\
Kheronighat & 0.5369054699071945 & 0.5658441262374312 & 0.028938656330236734 \\
Kibithu & 0.11911187001776047 & 0.13034886582331529 & 0.011236995805554817 \\
Manas NH Crossing & 0.29809200942195624 & 0.3196220545834817 & 0.021530045161525457 \\
Margherita & 0.17768562066267957 & 0.16697132710264284 & -0.010714293560036725 \\
Mathanguri & 0.15863617688268716 & 0.16311326711317198 & 0.004477090230484815 \\
Matunga & 0.050344635450549124 & 0.09791442734420186 & 0.047569791893652735 \\
Naharkatia & 0.3165976242146106 & 0.3072153402238076 & -0.009382283990802986 \\
Nanglamoraghat & 0.5855394105846985 & 0.5774178406566732 & -0.008121569928025263 \\
Neamatighat & 0.4874680766092895 & 0.5963765192370596 & 0.10890844262777011 \\
Numaligarh & 0.5480987143354679 & 0.542264638302129 & -0.005834076033338853 \\
Pagladiya N.T.Road X-ING & 0.20836194390983376 & 0.24480898735903844 & 0.036447043449204686 \\
Panbari & 0.24456172629159956 & 0.2449293515799722 & 0.00036762528837264163 \\
Passighat & 0.059395888047660605 & 0.15037602478579704 & 0.09098013673813643 \\
Puthimari N.H X-ING & 0.25037340454982393 & 0.28208453691665725 & 0.03171113236683332 \\
Seppa & 0.11646708708369591 & 0.1445160785925852 & 0.028048991508889287 \\
Sivasagar & 0.35968206910588096 & 0.36774523888722543 & 0.00806316978134447 \\
Suklai & 0.23730410975983496 & 0.24127796559630577 & 0.003973855836470808 \\
Tezpur & 0.8415065285695466 & 0.7858503277945921 & -0.05565620077495448 \\
Tezu & 0.11770004892932584 & 0.10769783158601853 & -0.010002217343307307 \\
        \bottomrule
        \end{tabular}
        &
        \begin{tabular}[t]{l n{1}{3} n{1}{3} n{1}{3}}
            \multicolumn{4}{c}{Ganga} \\
            \toprule
            \text{Basin Name} & \text{Linear} & \text{HydroNets} & \text{Diff} \\
            \midrule
Auralya & 0.2924280872048822 & 0.30472394072224596 & 0.01229585351736373 \\
Banda & 0.6705910336918739 & 0.6836292384335845 & 0.01303820474171058 \\
Chillaghat & 0.5520156264832755 & 0.7347791852147051 & 0.18276355873142958 \\
Etawah & 0.3199532511938191 & 0.34887872136043463 & 0.028925470166615508 \\
Gaisabad & 0.44250679775838864 & 0.48224566417641745 & 0.03973886641802882 \\
Garrauli & 0.49479745081499416 & 0.5193523121468859 & 0.024554861331891775 \\
Hamirpur & 0.17919800461891522 & 0.1771406671326362 & -0.0020573374862790095 \\
Kalpi & 0.37933718671550687 & 0.4201586180825567 & 0.040821431367049854 \\
Kora & -0.5415282761182365 & -0.052580157128803595 & 0.4889481189894329 \\
Madla & 0.45782183230134577 & 0.495423753617532 & 0.037601921316186226 \\
Nautghat & 0.30365645039633693 & 0.34756461544492745 & 0.04390816504859052 \\
Shahijina & 0.4823139641158328 & 0.476538691092261 & -0.005775273023571836 \\
Udi & 0.3991852629166368 & 0.3974146891071355 & -0.0017705738095012968 \\
            \bottomrule \\
            \multicolumn{4}{c}{Brahmaputra Improvement (Diff)} \\
            \multicolumn{4}{c}{
                \includegraphics[height=13em]{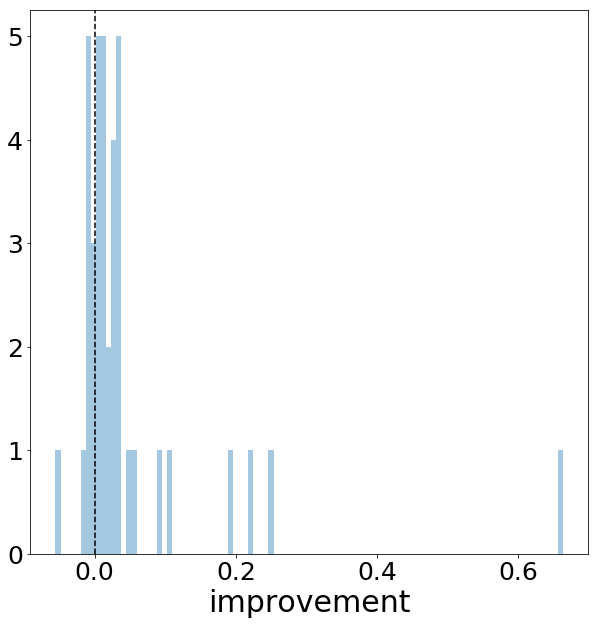}
            } \\ 
            \multicolumn{4}{c}{Ganga Improvement (Diff)} \\
            \multicolumn{4}{c}{
                \includegraphics[height=13em]{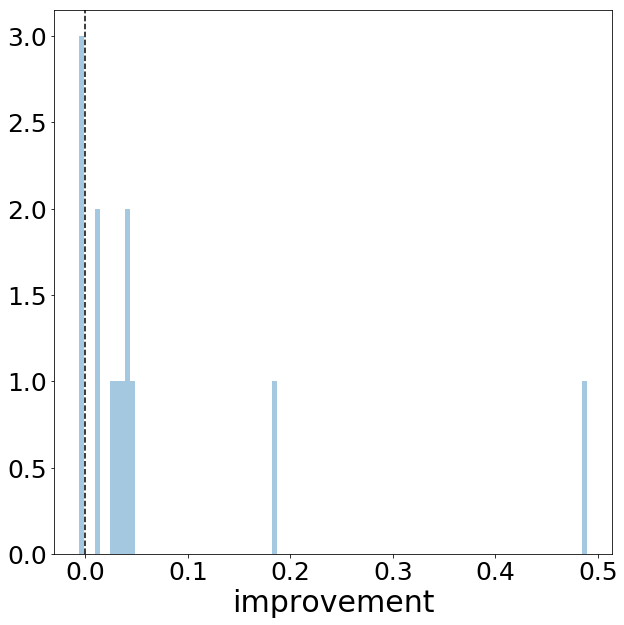}
            }
        \end{tabular} 
    \end{tabular}
    \caption{Full results over all basins in the Brahmaputra and Ganga regions.}
    \label{table:all_basins_results}
\end{table}

\newpage

\section {$R^2$ results}
\label{appendix:r2_results}

In Figure~\ref{fig:DescendantsDepth} and Figure~\ref{fig:MainComparison} we presented the results of Experiment 1 using the $R^2$-persist metric. Following are the $R^2$ results for the same experiment.

\vspace{1em}
\begin{figure}[h]
    \centering
    \begin{tabular}{cc}
        Brahma & Ganga \\
        \includegraphics[width=0.5\textwidth]{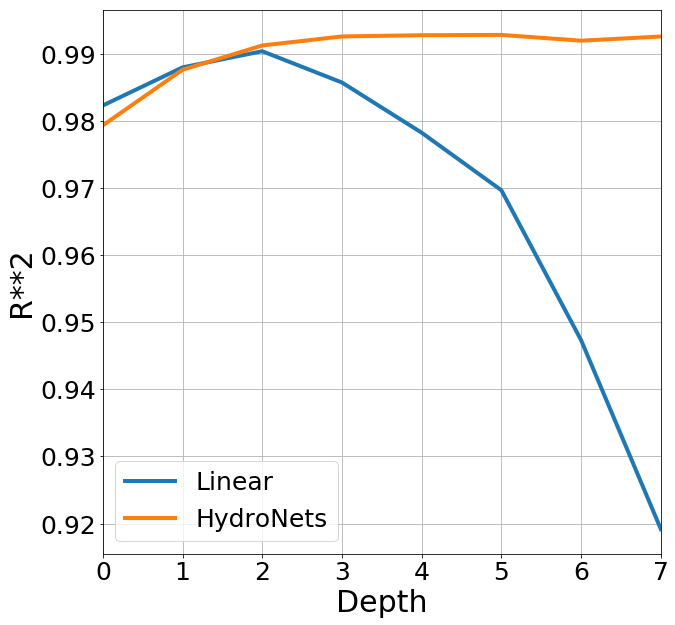}
        &
        \includegraphics[width=0.5\textwidth]{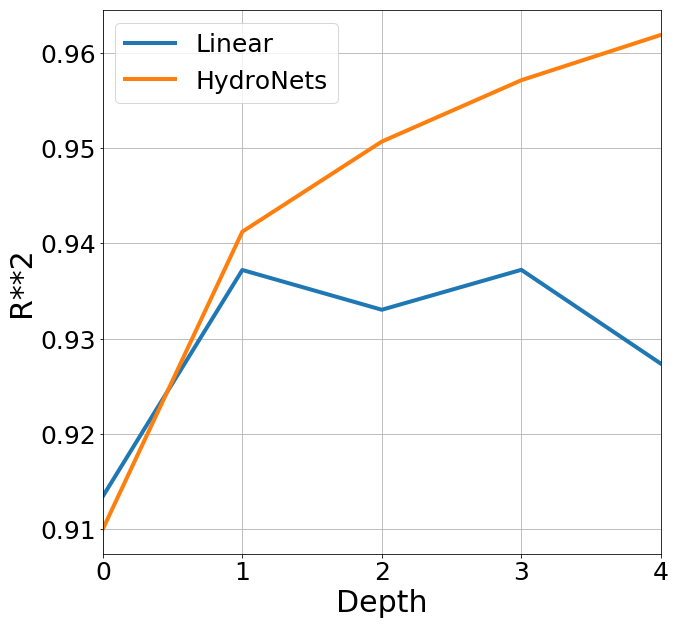}
        \\
        \includegraphics[width=0.5\textwidth]{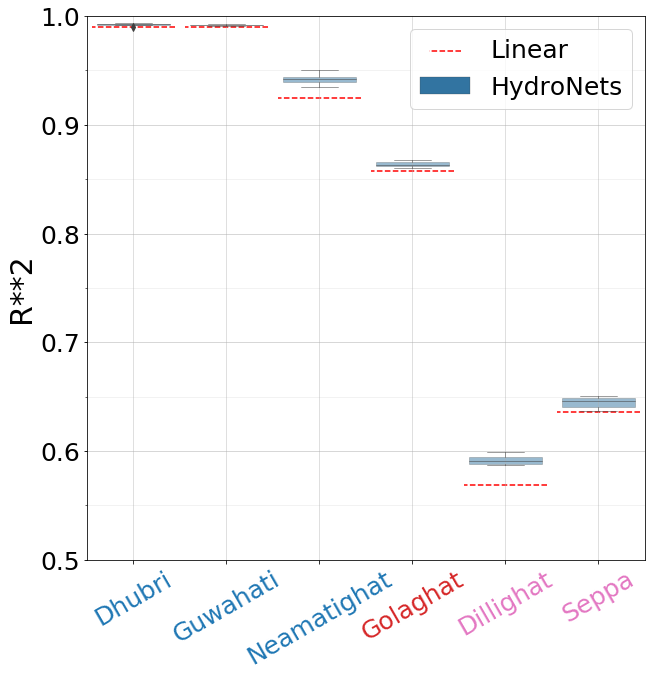}
        &
        \includegraphics[width=0.5\textwidth]{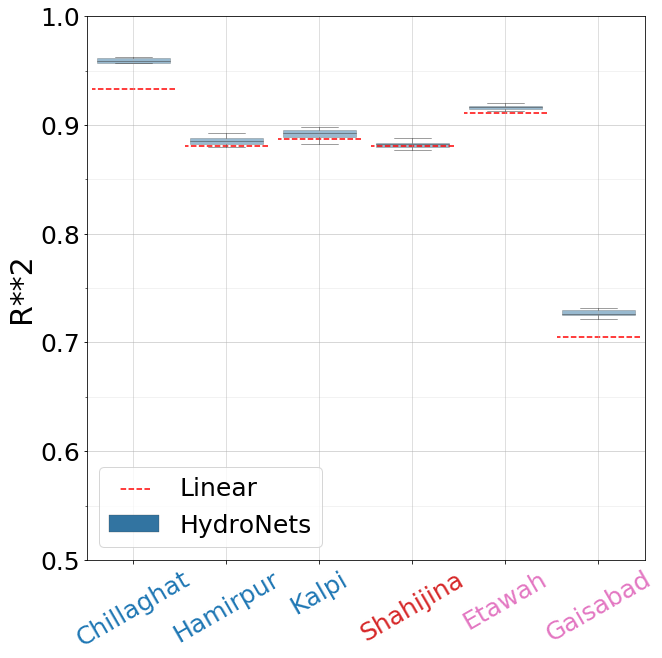} 
    \end{tabular}
    \caption{$R^2$ results}
\end{figure}

\end{document}